\documentclass[conference]{IEEEtran}

\usepackage{cite}
\usepackage{amsmath,amssymb,amsfonts}
\usepackage{algorithmic}
\usepackage{graphicx,color,subfigure}
\usepackage{textcomp}
\usepackage{booktabs} 
\usepackage{array} 
\usepackage{paralist} 
\usepackage{enumitem}
\usepackage{verbatim} 
\usepackage[linesnumbered,ruled]{algorithm2e} 
\usepackage{tabularx}  
\usepackage{engord}
\usepackage{booktabs,multirow}
\usepackage{url}
\usepackage{comment}
\usepackage{nomencl}

\usepackage{float}
\usepackage{mathtools}
\usepackage{bbding}
\usepackage{amsthm}
\usepackage[T1]{fontenc}

\begin{document}
%
\title{M-SET: Multi-Drone Swarm Intelligence Experimentation with Collision Avoidance Realism}

\author{
\IEEEauthorblockN{Chuhao Qin}
\IEEEauthorblockA{School of Computing\\
University of Leeds\\
Leeds, UK\\
sccq@leeds.ac.uk}
\and
\IEEEauthorblockN{Alexander Robins}
\IEEEauthorblockA{School of Computing\\
University of Leeds\\
Leeds, UK\\
alexjosephrobins@gmail.com}
\and
\IEEEauthorblockN{Callum Lillywhite-Roake}
\IEEEauthorblockA{School of Computing\\
University of Leeds\\
Leeds, UK\\
callum.methold@sky.com}
\and
\IEEEauthorblockN{Adam Pearce}
\IEEEauthorblockA{School of Computing\\
University of Leeds\\
Leeds, UK\\
adam.pearce2000@gmail.com}
\and
\IEEEauthorblockN{Hritik Mehta}
\IEEEauthorblockA{School of Computing\\
University of Leeds\\
Leeds, UK\\
hritikmehta10@gmail.com}
\and
\IEEEauthorblockN{Scott James}
\IEEEauthorblockA{School of Computing\\
University of Leeds\\
Leeds, UK\\
scottowenjames33@gmail.com}
\and
\IEEEauthorblockN{Tsz Ho Wong}
\IEEEauthorblockA{School of Computing\\
University of Leeds\\
Leeds, UK\\
jerrywongtszho@gmail.com}
\and
\IEEEauthorblockN{Evangelos Pournaras}
\IEEEauthorblockA{School of Computing\\
University of Leeds\\
Leeds, UK\\
E.Pournaras@leeds.ac.uk}
}

\maketitle

\begin{abstract}
Distributed sensing by cooperative drone swarms is crucial for several Smart City applications, such as traffic monitoring and disaster response. Using an indoor lab with inexpensive drones, a testbed supports complex and ambitious studies on these systems while maintaining low cost, rigor, and external validity. This paper introduces the Multi-drone Sensing Experimentation Testbed (M-SET), a novel platform designed to prototype, develop, test, and evaluate distributed sensing with swarm intelligence. M-SET addresses the limitations of existing testbeds that fail to emulate collisions, thus lacking realism in outdoor environments. By integrating a collision avoidance method based on a potential field algorithm, M-SET ensures collision-free navigation and sensing, further optimized via a multi-agent collective learning algorithm. Extensive evaluation demonstrates accurate energy consumption estimation and a low risk of collisions, providing a robust proof-of-concept. New insights show that M-SET has significant potential to support ambitious research with minimal cost, simplicity, and high sensing quality.
\end{abstract}


%
\IEEEpeerreviewmaketitle

\section{Introduction}
Unmanned Aerial Vehicles (UAVs), or drones, can form swarms to improve collaboration and efficiency in sensor data collection for Smart Cities, such as early traffic congestion reporting and natural disaster mapping~\cite{butilua2022urban}. Using a single high-profile drone is costly and limited by flight range. In contrast, multiple low-cost cooperative drones are versatile to cover wider areas and allow for battery recharging. To benefit from this flexibility, drones require coordinated actions with autonomy and computational intelligence, which can be solved by distributed optimization and multi-agent learning algorithms, such as swarm intelligence~\cite{zhou2020uav}. To prototype, develop, test and evaluate distributed sensing exhibiting swarm intelligence, earlier testbeds have been introduced~\cite{ding2023distributed,qin20223}. Using an indoor lab with cheap small drones, they emulate complex outdoor sensing environments without the need for costly equipment, regulations, or environmental uncertainties.

However, these existing indoor testbeds lack the realism of outdoor environments, particularly regarding in-flight safety issues such as obstacle-to-drone and drone-to-drone collisions. Designing and prototyping a testbed that integrates collision avoidance in swarm intelligence for distributed sensing is a complex challenge. On the one hand, swarm intelligence enables autonomous and flexible coordination for efficient sensing missions, but increases the risk of unpredictable collisions, requiring highly sophisticated algorithms for collision detection and avoidance, especially in small lab spaces. On the other hand, incorporating collision avoidance introduces new hardware testing challenges, as it alters the navigation and sensing outputs by swarm intelligence, impacting factors such as energy consumption and sensing accuracy.

To tackle these challenges, this paper introduces the Multi-drone Sensing Experimentation Testbed (M-SET), designed to improve the realism of multi-drone sensing operations. This testbed studies various distributed sensing problems in drones, including charging control, navigation, and collision avoidance. As a proof-of-concept, an artificial potential field algorithm~\cite{fedele2018obstacles} is applied to predict collision fields and determine optimal drone flight paths, effectively detecting and mitigating potential collisions. This method ensures collision-free navigation and sensing, which are further coordinated and optimized using a multi-agent collective learning approach~\cite{pournaras2018decentralized,pournaras2020collective}. Extensive experimentation with low-cost real drones, a virtual sensing environment, and real-world traffic data from Athens~\cite{barmpounakis2020new} validates the effectiveness of M-SET in traffic vehicle monitoring, demonstrating its capacity to move complex swarm intelligence and collision avoidance algorithms for drones to real-world. 

The contributions of this paper are outlined as follows: (i)  A new and generic testbed model, M-SET, to enhance the realism of distributed sensing with drones exhibiting swarm intelligence by integrating collision avoidance method. (ii) A first functional prototype of this testbed with a proof-of-concept on accurate estimate of energy consumption and low risk of collisions. (iii) An environment of higher realism to test multi-agent collective learning~\cite{pournaras2018decentralized} for drone navigation and sensing, as well as artificial potential field~\cite{fedele2018obstacles} for collision avoidance, using real traffic monitoring data~\cite{barmpounakis2020new}. (iv) An open-source software platform\footnote{https://github.com/TDI-Lab/M-SET} and documentation\footnote{https://github.com/TDI-Lab/M-SET-Documentation} as a benchmark, providing detailed, reproducible coding examples and instructions for M-SET to foster future development and collaboration within the broader community. (v) A thorough evaluation of the prototype highlights new insights into the low cost, safety and applicability of drone sensing.

\begin{figure}
    \centering
    \includegraphics[width=0.8\linewidth]{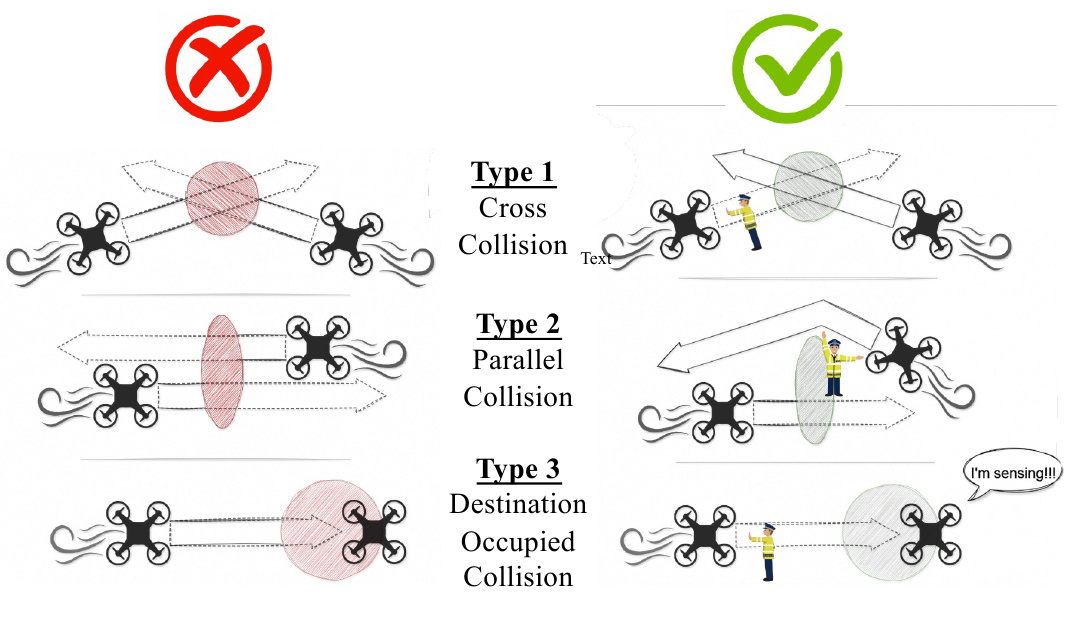}
    \caption{Three types of collisions and corresponding avoidance methods. Cross collision denotes two drones fly across each other; Parallel collision indicates two drones fly towards each other; Destination-occupied collision means one drone performing sensing occupies another drone's destination.}
    \label{fig:basic_ca}
\end{figure}

\begin{figure*}
    \centering
    \includegraphics[width=0.8\linewidth]{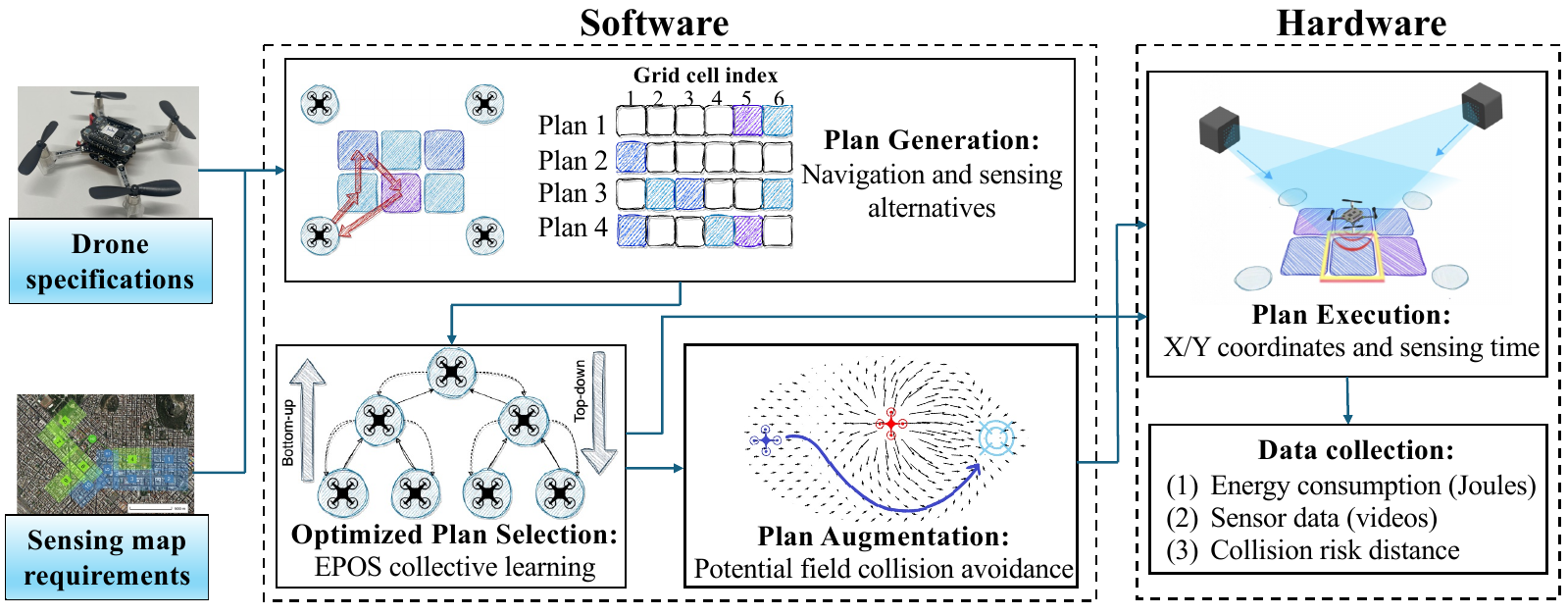}
    \caption{An overview of the prototyped M-SET architecture.}
    \label{fig:testbed}
\end{figure*}

\section{Related Work}
Distributed sensing with a swarm of autonomous and intelligent drones has been traditionally defined as an NP-hard combinatorial optimization problem~\cite{zhou2020uav}. Various distributed sensing solutions, such as particle swarm optimization~\cite{gao2018multi}, flocking~\cite{dai2019swarm}, and wolf-pack search~\cite{chen2018multi}, have been applied to coordinate the actions of drones swarms and assign their sensing tasks. To validate realism and external validity of these algorithms, previous work has built testbed for real-world implementation and assessment. Khan et al.~\cite{khan2017mobile} tests the feasibility of mobile target tracking algorithms using clustering and cover-set coverage methods with help of Parrot AR.Drone quadcopters. The testbed in~\cite{bouachir2019testbed} demonstrates the performance of a dedicated Quality-of-Service communication system using Paparazzi drones for cooperative sensing missions. However, these testbeds fail to emulate collisions with fixed or dynamic obstacles, leading to inaccuracies in analysis of multi-drone operations in real-world scenarios. 

Collision avoidance algorithms find applicability to distributed sensing with drones in order to facilitate safe and reliable airspace access~\cite{rezaee2024comprehensive}. Computer vision and machine learning-based techniques that utilise a combination of sensors such as LiDAR, Radar and Sonar propose onboard collision avoidance strategies~\cite{rezaee2024comprehensive,kouris2018learning}. Artificial potential fields is another widely known collision avoidance strategy, commonly used in robotics~\cite{fedele2018obstacles}. Batinovic et al.~\cite{batinovic2022path} uses artificial potential field for obstacle avoidance in 3D Environments by equipping drones with LiDAR sensors, and eliminated the possibility of agents getting stuck in a local minima employing rotational components of the repulsive force. Nevertheless, they only conduct experiments in simulation, lacking the demonstration in applicability and realism. Sabikan et al.~\cite{sabikan2020modelling} addressed this issue by developing a Time-to-Collision mathematical model using particle swarm optimization to find collision-free paths for outdoor drone data recording, but their work is limited to a platform with a single quadcopter. Schmittle et al.~\cite{schmittle2018openuav} establishes an open-source, cloud-enabled testbed to study the navigation around obstacles and drone swarm formation. However, it is limited to an exclusive software-based simulation and does not analyze how collision avoidance impacts the assignment of sensing tasks to drones, particularly regarding energy consumption and sensor data collection.

The designed testbed overcome these barriers by integrating potential field algorithm for collision avoidance with our previous testbed~\cite{qin20223}. This enhanced testbed can adapt to various application scenarios and contexts while ensuring safety during navigation and sensing. By enhancing realism, it effectively emulate the outdoor environments. Furthermore, it examines the impact of collision avoidance on energy consumption and data collection, offering new insights into hardware testing for drones.

\section{Testbed Design}
M-SET relies on a model, which can be implemented in different lab environments. This paper focuses on the sensing missions of traffic monitoring using camera-equipped drones for city road surveillance. The primary objective is to manage the spatio-temporal flight behaviors of a drone swarm to accurately observe vehicles and traffic flow, aiding in the detection of congestion and accidents. At an abstract level, the testbed is modeled by the elements presented in the rest of this section. 

\subsection{Testbed components}

\subsubsection{UAVs}
They communicate to interact with each other directly, or via low-latency edge proxies, or through the cloud~\cite{nezami2023smotec}. Each drone can run its swarm intelligence and collision avoidance software for distributed sensing within the following continuum~\cite{fanitabasi2020self}: (i) offline/online, remote, centralized computations (server deployment), (ii) offline/online, remote, distributed computations (edge-to-cloud deployment scenario)~\cite{nezami2023smotec}, and (iii) online, locally on drones, distributed computations. For long-term sensing missions, each drones can support wireless charging and be fully charged before starting the next missions~\cite{chittoor2021review}.

\subsubsection{Virtual sensing environment}
It is a video, displayed by a monitor or a projector, where drones sense at various resolutions based on their flying or hovering altitude and screen resolution. Each altitude corresponds to a specific size of a grid cell based on the field of view that the camera of a drone has, affecting recorded image resolution. Different areas have distinct sensing requirements that determine drone hovering time and data collection. For instance, high-traffic zones have higher sensing requirement (i.e., longer hovering time) for accurate traffic flow observation.

\subsubsection{Swarm intelligence}
It assists multiple drones to plan the navigation and sensing in a coordinated way such that each self-select one plan influenced by the selections of others. Therefore, the total sensing by the swarm matches well the sensing requirements of all cells. This matching represents the relative approximation between the total actual sensed values per cell and sensing requirements per cell. Error and correlation metrics such as the root mean squared error, cross-correlation or residuals of summed squares can estimate this matching~\cite{pournaras2020collective}. 

\subsubsection{Collision detection and collision avoidance}
It detects all possible in-flight and static collisions, i.e., intersections of flight paths or walls in multi-drone missions, and then minimizes the likelihood of collisions during the path planning. This requires an intelligent path planning algorithm, making distributed sensing cost-effective and safe within various scenarios. The algorithm detects intersections of flight paths and obstacles, and augments the navigation and sensing plans selected via swarm intelligence to prevent the potential collisions, minimizing in-flight risks, i.e., the traveling distance at high risk of collisions. Fig~\ref{fig:basic_ca} illustrates three typical types of collisions considered in this paper (cross, parallel, and destination-occupied).

\subsection{Architecture overview}\label{sec:overview}
Fig.~\ref{fig:testbed} illustrates an overview of M-SET architecture. The core of M-SET lies in two software approaches. One is the decentralized multi-agent collective learning method, named \emph{Economic Planning and Optimized Selections} (EPOS)~\cite{pournaras2018decentralized,pournaras2020collective,qin2023coordination}. It generates for each agent a finite number of discrete navigation and sensing options, each with an estimated power consumption: the \emph{possible plans} and their \emph{cost} respectively. Plan generation is performed using the drones specifications (weight, propeller and battery parameters), and the sensing environment (wind and grid cells). Then the agents interact iteratively in a bottom-up and top-down fashion over a tree communication structure to make a selection such that all choices together add up to maximize the sensing quality (matching). However, these selected plans have no information about the potential collisions, and thus the path planning algorithm for collision avoidance is required to augment selected plans. This testbed uses the artificial potential field algorithm~\cite{fedele2018obstacles} to generate attractive forces towards the destinations and repulsive forces between drones. These forces guide the drones to their target cells for sensing according to the selected plan, while also avoiding potential collisions. Finally, the algorithm produces the output plans containing the X/Y coordinates and sensing (hovering) duration at each cell. They are then executed by drones in hardware, which records the energy consumption, the collected sensor data and collision risk distance during the sensing mission. Therein, collision risk distance implies the specific distance at which the risk of collisions becomes significant, illustrated by the path length within the shaded area shown in Fig.~\ref{fig:basic_ca}.

\section{Testbed Prototyping}
This section introduces the prototyping details of M-SET architecture.

\begin{figure}
	\centering
 	\subfigure[Crazyflie for testing positioning and wireless charging.]{
		\includegraphics[width=0.4\linewidth]{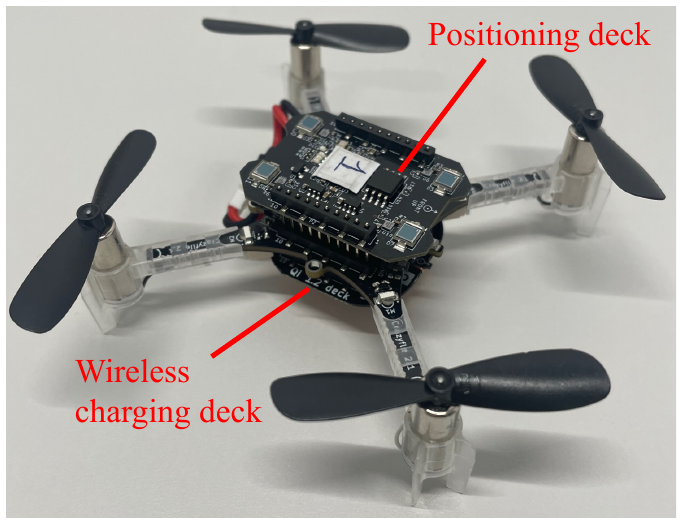}
		\label{fig:drone_charge}
	}
 \hspace{0.1cm}
	\subfigure[Crazyflie for navigation and camera recording.]{
		\includegraphics[width=0.4\linewidth]{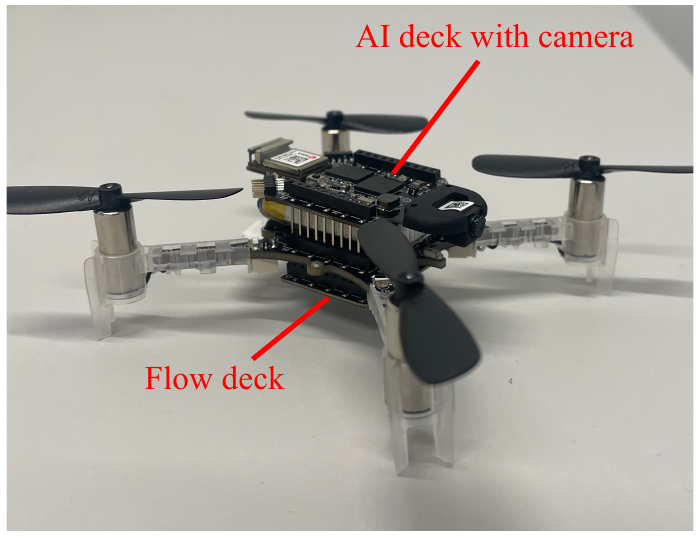}
		\label{fig:drone_camera}
	}
 	\caption{Assembly of Crazyflies for two types of functions.}
	\label{fig:sensing_map}
\end{figure}

\subsection{Hardware UAVs for sensing}
M-SET uses the Crazyflie 2.1 because of its size, weight and accessibility\footnote{https://www.bitcraze.io/products/crazyflie-2-1/}. The drone has 27g weight, 47mm propeller length, a battery capacity of 250mAh (LiPo battery) and it is configured to fly with an average ground speed of 0.1m/s. Based on this information, the power consumption can be estimated. Crazyflie can be programmed in Python with the support of API in Bitcraze\footnote{https://github.com/bitcraze/crazyflie-lib-python}. This is further expanded by the Crazyswarm API\footnote{https://github.com/USC-ACTLab/crazyswarm}, which provides additional functionality to control a swarm of Crazyflies. The drone can be mounted by multiple hardware decks to support different functions, such as positioning (with the support of lighthouse base stations), camera recording, and wireless charging. Due to maximum weight limitation of Crazyflie, two types of drones are assembled, one is for testing wireless charging, see Fig.~\ref{fig:drone_charge}, and the other is for collecting sensor data, see Fig.~\ref{fig:drone_camera}. The data-collection drone with an ultra low power 320×320 grayscale camera is equipped with a tiny mirror to take videos of ground.

\begin{figure}
	\centering
 	\subfigure[A 75 inch screen to display the traffic flow of vehicles.]{
		\includegraphics[width=0.45\linewidth]{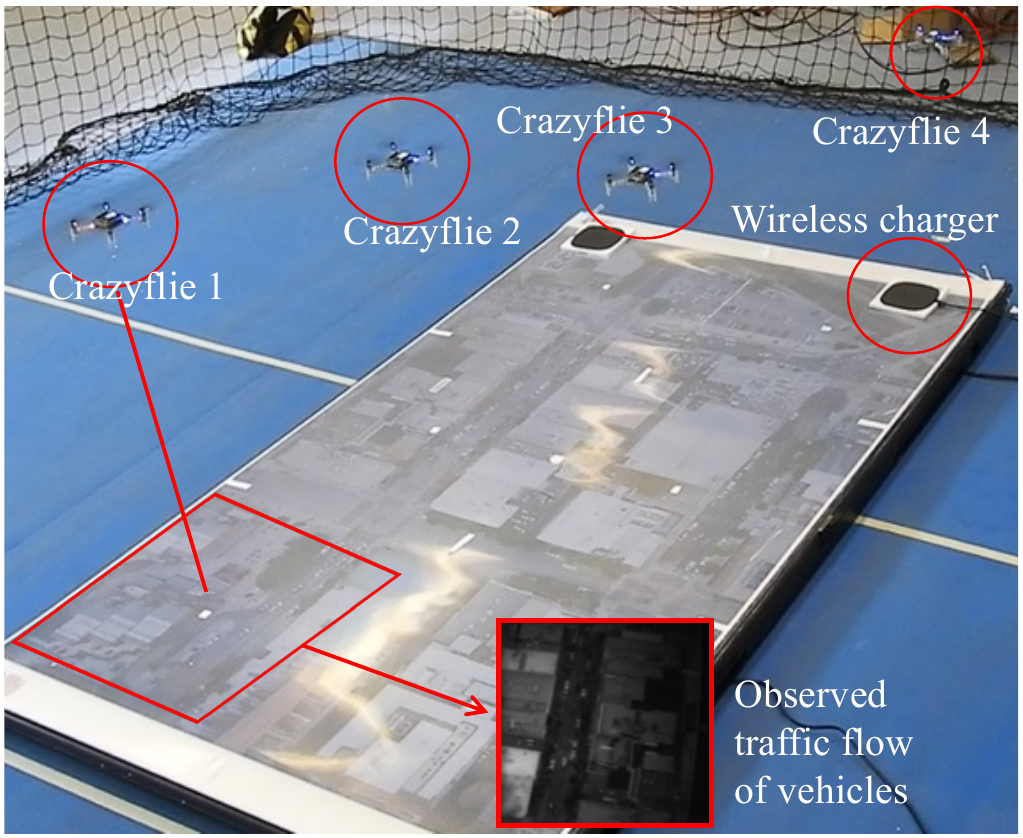}
		\label{fig:screen}
	}
 \hspace{0.1cm}
	\subfigure[Grid cells in the video and sensing requirements.]{
		\includegraphics[width=0.45\linewidth]{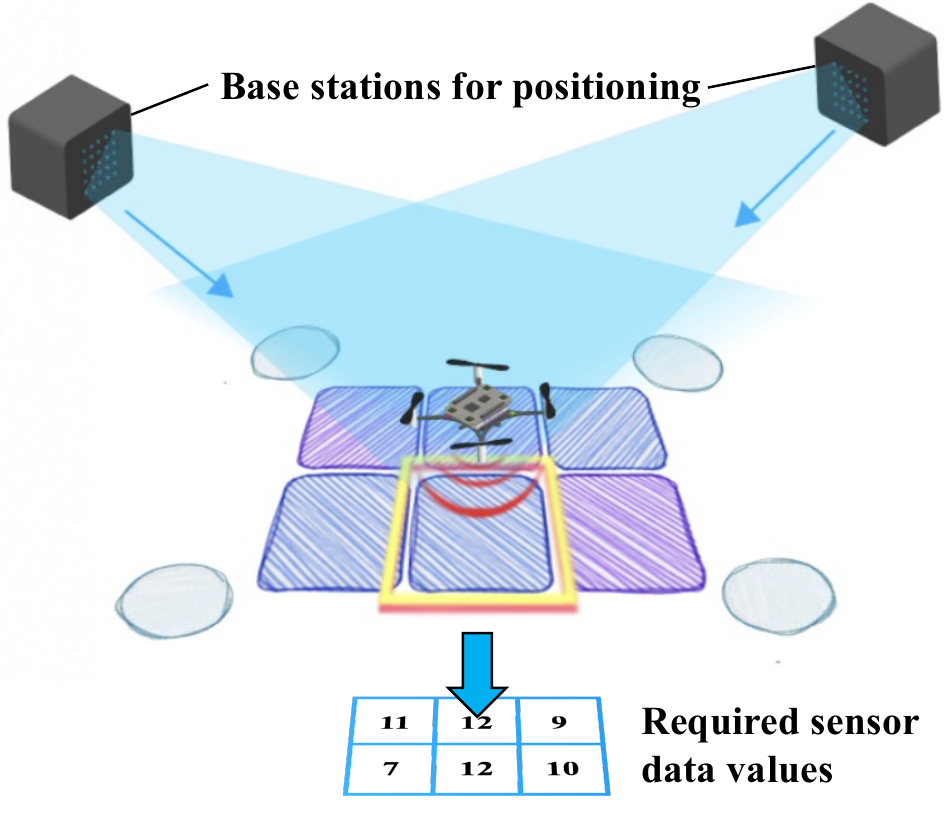}
		\label{fig:sensing_requirement}
	}
 	\caption{Indoor sensing lab using a large screen, Crazyflies, wireless chargers, and lighthouse base stations for positioning. Drones fly to grid cells and collect sensor data by recording the videos of traffic vehicles. }
	\label{fig:sensing_map}
\end{figure}

\subsection{Setting up an indoor sensing environment}
A 75 inch screen is set on the ground as an indoor environment to emulate the outdoor sensing environments. As shown in Fig.~\ref{fig:screen} and Fig.~\ref{fig:sensing_requirement}, the screen is divided into 2x3 square grid cells, each with an area of 55x47cm. Each cell can be sensed effectively from an altitude of 50cm (i.e., the vehicles can be clearly observed). Other accessory equipment is set, including lighthouse base stations and wireless chargers. To show the significant and broad impact of M-SET on a transportation scenario, the screen displays the video of the traffic flow of vehicles recorded from satellites or other simulators, e.g., Simulations of Urban Mobility (SUMO). In this scenario, a higher sensing requirement in a cell represents a high urgency for traffic monitoring over this area, which requires drones to hover a longer time to measure accurately. M-SET uses a real-world dataset of vehicle trajectories named pNEUMA\footnote{https://open-traffic.epfl.ch/} collected by a swarm of drones in the congested downtown area of Athens, Greece~\cite{barmpounakis2020new}. 

\subsection{Swarm intelligence using collective learning}
A number of $N$ agents, each corresponding to a drone, autonomously generate 16 plans, each comprising sequences of $M$ real values representing sensing duration at cells. These plans, serving as random samples of alternative routes with a random number of cells, respect battery constraints. They are assigned costs based on the power consumption of flying and hovering~\cite{stolaroff2018energy}. 

The plan selection is made in a coordinated way using EPOS\footnote{https://github.com/epournaras/epos}~\cite{pournaras2018decentralized} to prevent over-sensing and under-sensing\footnote{Over-sensing causes excessive data that needs further processing, waste of energy consumption, high storage and privacy cost~\cite{pournaras2024collective}, while under-sensing fails to satisfy sensing requirements.}. The algorithm is selected because of its remarkable scalability (support a large number of agents), efficiency (low communication and computational cost) and resilience~\cite{pournaras2018decentralized, pournaras2020collective,majumdar2023discrete}. For this, agents connect into a balanced binary tree topology within which they interact with their children and parent to improve iteratively their plan selection. The goal of the agents is to minimize the residual of sum squares (RSS) between the following unit-length scaled signals: sensed values per cell summed up over all agents and the sensing requirements per cell. The agents perform $40$ bottom-up and top-down learning iterations. More information about EPOS is out of the scope of this paper and can be found in earlier work~\cite{pournaras2018decentralized, pournaras2020collective}. For this testbed prototype, the optimization process is performed offline and remotely, however deployments of EPOS for online optimization are already available for future extensions~\cite{fanitabasi2020self}. 

\begin{figure}
    \centering
 	\subfigure[Layout of potential field grids to the target drone.]{
		\includegraphics[width=0.4\linewidth]{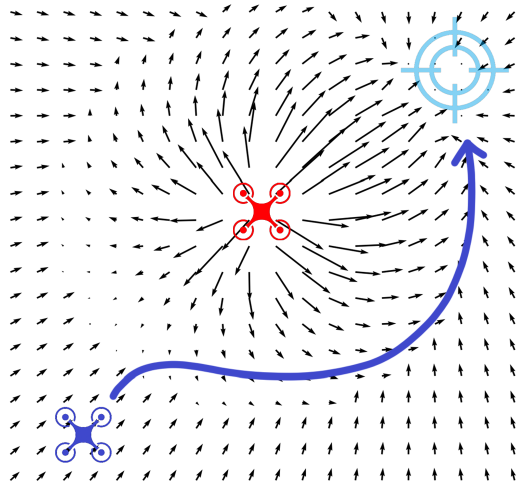}
		\label{fig:pfg_ex}
	}
 \hspace{0.1cm}
	\subfigure[An example of collision avoidance with 2 drones.]{
		\includegraphics[width=0.4\linewidth]{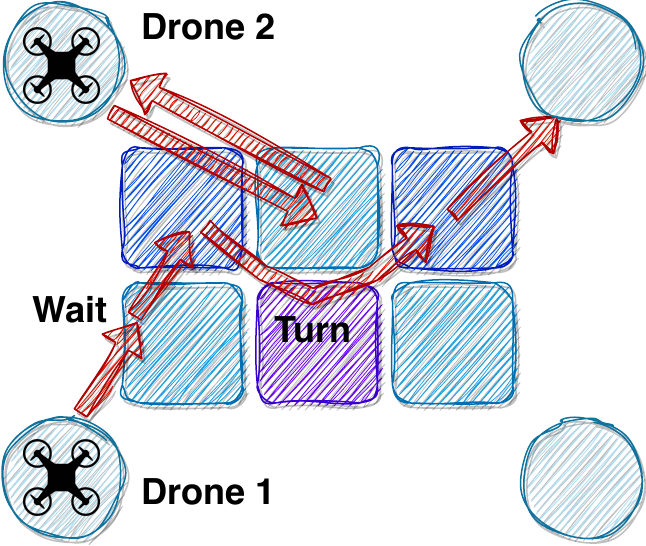}
		\label{fig:pf_ca_ex}
	}
    \caption{An example of collision avoidance using artificial potential field: (a) The blue drone is the target drone attracted by the destination, whereas the red drone is the obstacle drone repelling the target drone. The vectors influenced by both attractive and repulsive forces point towards the navigation of the target drone. (b) Since drone 2 has higher priority and stronger repulsive force than drone 1, drone 1 is pushed out of the target cell and "wait" until drone 2 passes by, and makes a "turn" when traveling to the next cell.}
    \label{fig:pfg}
\end{figure}

\subsection{Collision avoidance using artificial potential field}
M-SET applies an artificial potential field algorithm~\cite{fedele2018obstacles} to path planning of drones, which creates force to repel drones from obstacles and attract them towards their destinations. As shown in Fig.~\ref{fig:pfg}, a Potential Fields Grid (PFG) in this implementation is created for each drone (named as the target drone) per timestamp. It is a 2D-grid of vectors, where each vector points in the direction that the target drone should fly at that position per timestamp~\cite{fedele2018obstacles}. There are two types of PFG: attractive PFG and repulsive PFG, which coexist to balance the drone navigation. The attractive PFG generates forces that pull the drone towards its target destination. The repulsive PFG creates forces that push the drone away from obstacles (e.g., the other drones, walls and other static obstacles). These repulsive forces are stronger when the target drone is closer to an obstacle. After the summation of all types of forces, the drone navigates towards its destination while avoiding obstacles. Thus, given $I$ vectors, a vector with index of $i$ at timestamp $t$, $i \geq 1$, $t \geq 1$, consists of two components: attractive component $V^\mathsf{a}$ and repulsive component $V^\mathsf{r}$, formulated as follows:
\begin{equation}
    V_i (t) = \sum_{j \in \mathcal{O}} V^\mathsf{r}_{i,j} (t) + V^\mathsf{a}_i (t),
    \label{eq:vector}
\end{equation}
where the repulsive one is effected by the obstacle $j$, $j \in \mathcal{O}$, where $\mathcal{O}$ defines the set of obstacles in the map; the attractive component is influenced by the current destination for the target drone $n$ (this destination is changed once the drone reaches it). The attractive component can be formulated as:
\begin{equation}
V^\mathsf{a}_i(t) = V^\mathsf{a}_i(t-1).
\end{equation}

However, there is a the common problem with repulsive PFG: agents stand-off in situations where they continuously repel each other without making progress, leading to a deadlock scenario~\cite{fedele2018obstacles}. To mitigate the lock, the priorities of drones are assigned randomly. Drones with higher priority exert stronger and more extensive repulsive forces, pushing lower-priority drones out of their paths. As a result, drones can reach their respective destinations in sequence without getting obstructed by obstacles. The maximum radius of repulsion effect of a drone is defined as follows:
\begin{equation}
    R_j = D^\mathsf{min} (1 + ln P_j),
\end{equation}
where $P_j$ denotes the priority index of the obstacle $j$. The higher value of $P_j$ is, the higher priority of an obstacle drone is ($P_j$ is a constant if $j$ is a static obstacle like walls). $D^\mathsf{min}$ denotes the minimum distance between a drone and an obstacle before they collide. This paper sets 25cm considering the wind force caused by the Crazyfly. Thus, we can formulate the update of a vector in repulsive PFG $V^\mathsf{r}_{i,j}$ with index of $i$ per timestamp $t$ as follows:
\begin{equation}
\begin{split}
        V^\mathsf{r}_{i,j} (t) = \left \{
        \begin{array}{ll}
           \frac{V^\mathsf{r}_{i,j}(t-1) \cdot S_j^2}{|V^\mathsf{r}_{i,j}(t-1)| \cdot D_{i,j}(t)},  &  D_{i,j}(t) \leq R_j\\
           0,  &    otherwise
        \end{array},
        \right.
\end{split}
\label{eq:vectors_repulsive}
\end{equation}
\begin{equation}
    S_j = \delta |V^\mathsf{a}_i(t -1)| + ln P_j,
\end{equation}
where $D_{i,j}$ indicates the distance between the vector $i$ and an obstacle $j$; $S_j$ denotes the scaling factor for the strength of repulsion, higher than the strength (or magnitude) of attractive component of the vector $V^\mathsf{a}_i$. This ensures that the repulsive force acting on the target drone is stronger than attractive force, thus influencing the vector $V_i$ according to Eq.(\ref{eq:vector}). We set the scaling value $\delta = 2.5$ empirically to ensure that the lowest priority drone is strong enough to repel the other drones under the attractive forces. Therefore, the vectors in PFG prioritize maintaining a safe distance and prevent potential collisions over reaching the destinations. For normalization, we set the magnitude of both repulsive and attractive components as $|V^\mathsf{r}_i(1)| = |V^\mathsf{a}_i(1)|=1$. Finally, the target drone is forced by the summation of all vectors per timestamp, formulated as $\sum^I_{i=1} V_i (t)$ for $I$ vectors.

\section{Experimental Evaluation}
In this section, the baselines and performance evaluation metrics are illustrated. Evaluation are made using the testbed scenario and complex simulation scenarios.

\subsection{Baselines and metrics}
The approach used in M-SET is the collective learning of EPOS based on potential field collision avoidance, named as \emph{EPOS-PF}. To assess the collision avoidance, two baseline approaches are introduced: collective learning without collision avoidance (\emph{EPOS}) and collective learning with custom collision-based scheduler (\emph{EPOS-CA}). \emph{EPOS-CA} considers three classical types of collisions during the in-flight missions of drones as shown in Fig.~\ref{fig:basic_ca}. These collisions are detected after sensing plans are selected and avoided by delaying drones with lower priority. Cross and destination-occupied collisions are mitigated by controlling drones to wait until the path is collision-free while parallel collision is removed by redirecting drones to a point away from its original path before it resumes back to its target cell. To compare with \emph{EPOS} that minimizes RSS to improve sensing quality regardless of energy consumption, this paper introduces another baseline method that agents can make a choice that minimizes the energy consumption of their drones while preventing collisions using artificial potential field, named as \emph{Greedy-PF}.

The evaluation of all approaches includes key metrics:
\subsubsection{Energy consumption}
It is the total energy consumed by all drones, calculated by their hovering and traveling time.
\subsubsection{Risk of collisions}
It represents the ratio of the total travel distance where drones are at risk of collision. It can be calculated by $d_r / d$, where $d$ denotes the total traveling distance of drones, $d_r$ indicates the collision risk distance.
\subsubsection{Sensing mismatch}
It denotes the RSS between the total sensing of drones (i.e., the number of observed vehicles) per cell and the sensing requirements (i.e., the number of vehicles acquired from pNEUMA~\cite{barmpounakis2020new}) per cell.

\begin{figure}
	\centering
      \subfigure[Actual voltage logging of Crazyflies.]{
		\includegraphics[width=0.42\linewidth]{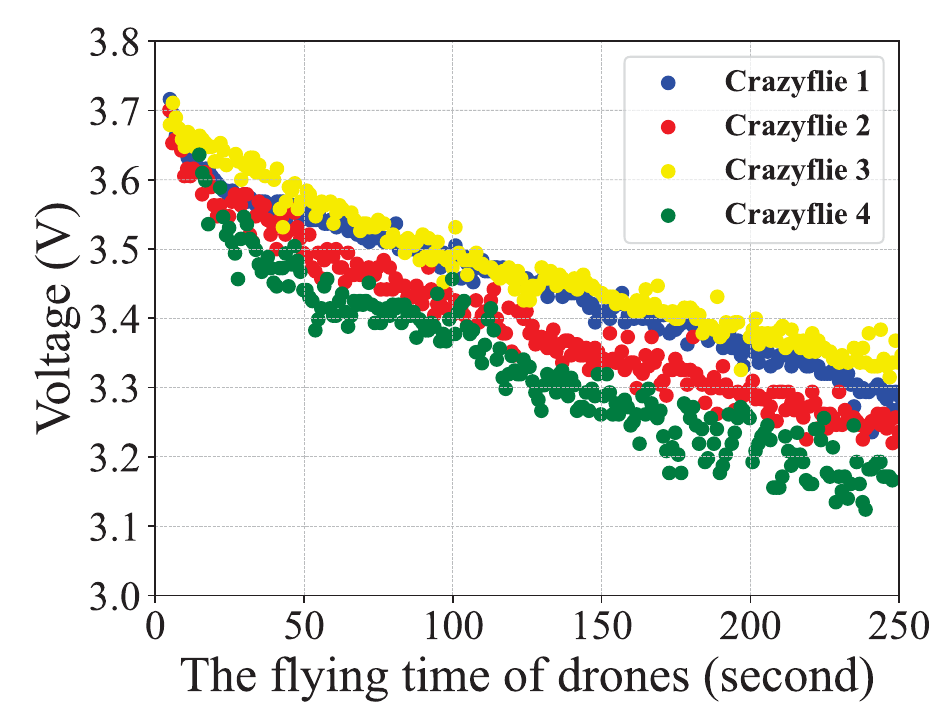}
		\label{fig:cf_voltage}
	}
    \hspace{0.2cm}
  	\subfigure[Real vs. Estimated energy consumption of Crazyflies.]{
		\includegraphics[width=0.42\linewidth]{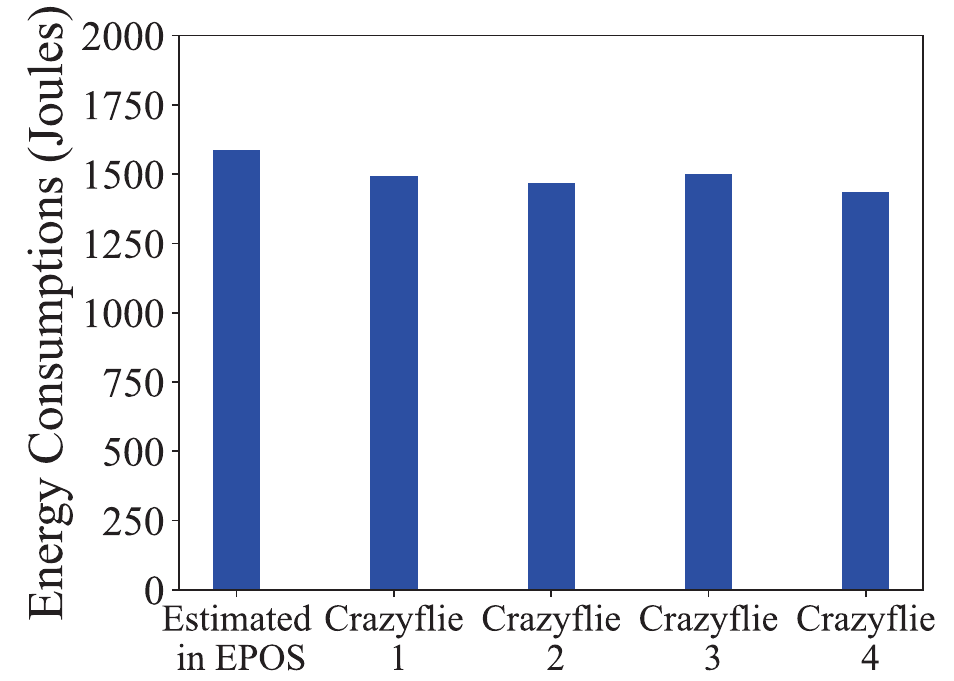}
		\label{fig:cf_energy}
	}
 	\caption{Comparison between estimated and real energy consumption. \textbf{M-SET has highly accurate energy consumption estimation in real-world.}}
	\label{fig:testbed_energy}
\end{figure}

\begin{figure}
	\centering
	\subfigure[Total energy consumption of drones in different approaches.]{
		\includegraphics[width=0.42\linewidth]{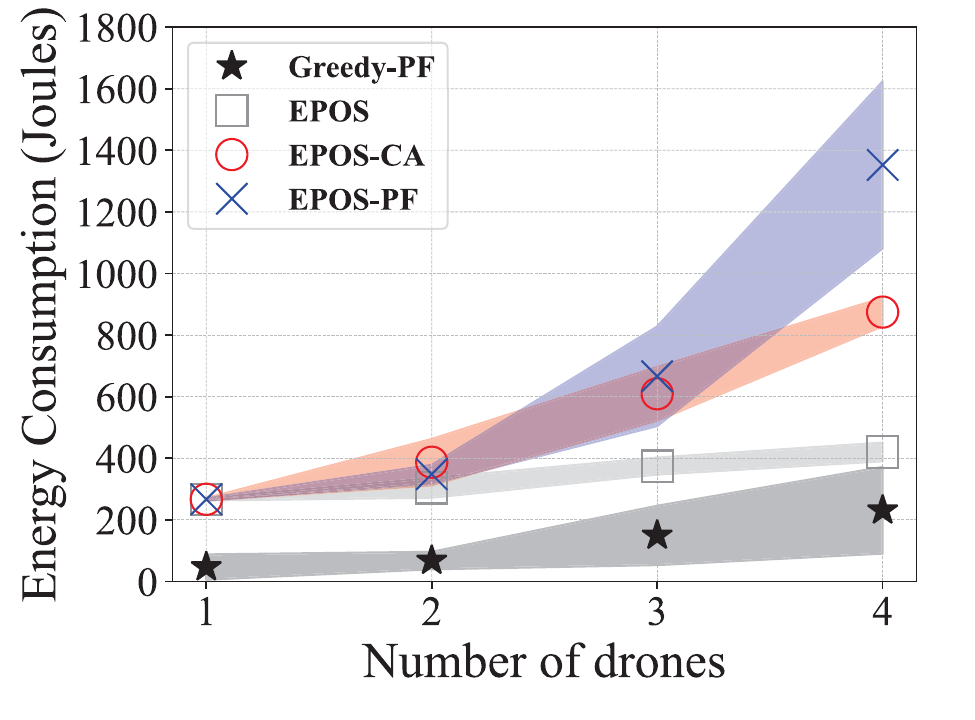}
		\label{fig:eval_energy}
	}
    \hspace{0.2cm}
 	\subfigure[Risk of collisions.]{
		\includegraphics[width=0.42\linewidth]{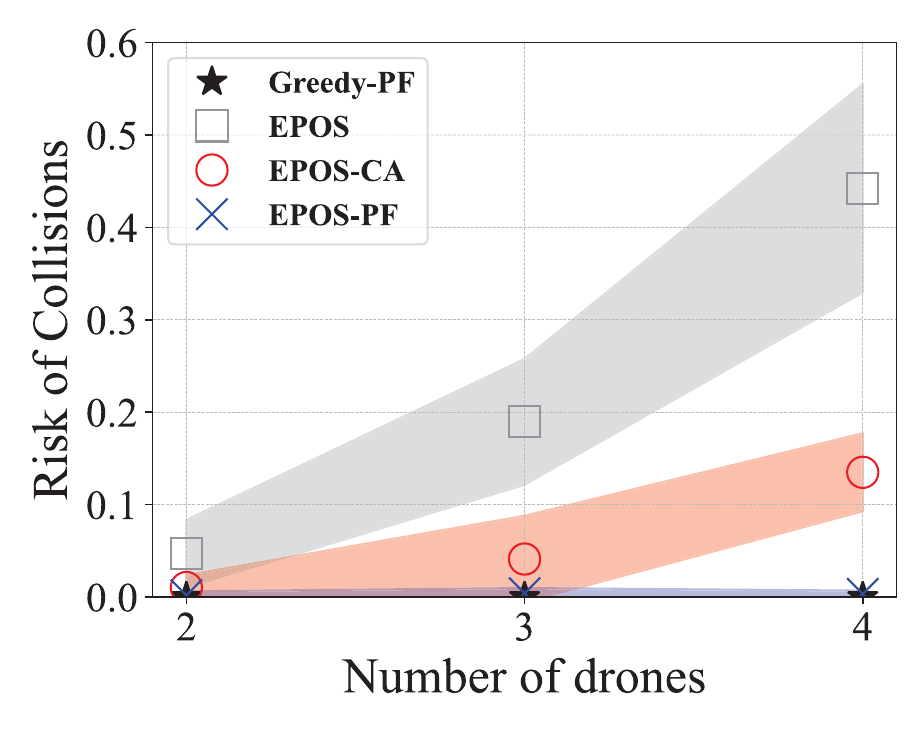}
		\label{fig:eval_risk}
	}
     \hspace{0.2cm}
  	\subfigure[Count of different types of collisions.]{
		\includegraphics[width=0.42\linewidth]{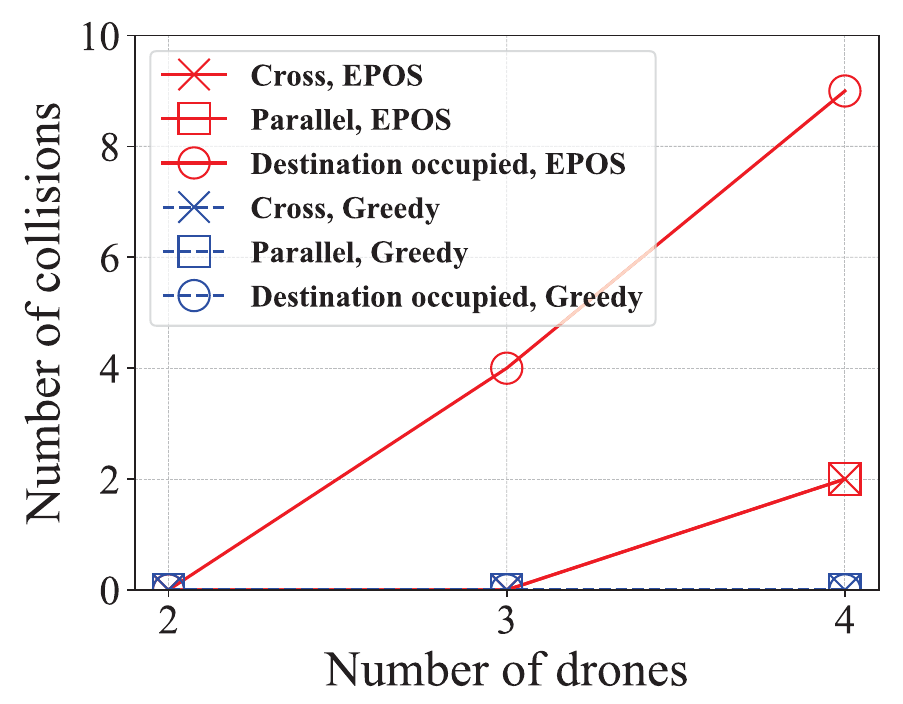}
		\label{fig:eval_collision_types}
	}
     \hspace{0.2cm}
	\subfigure[Sensing mismatch.]{
		\includegraphics[width=0.42\linewidth]{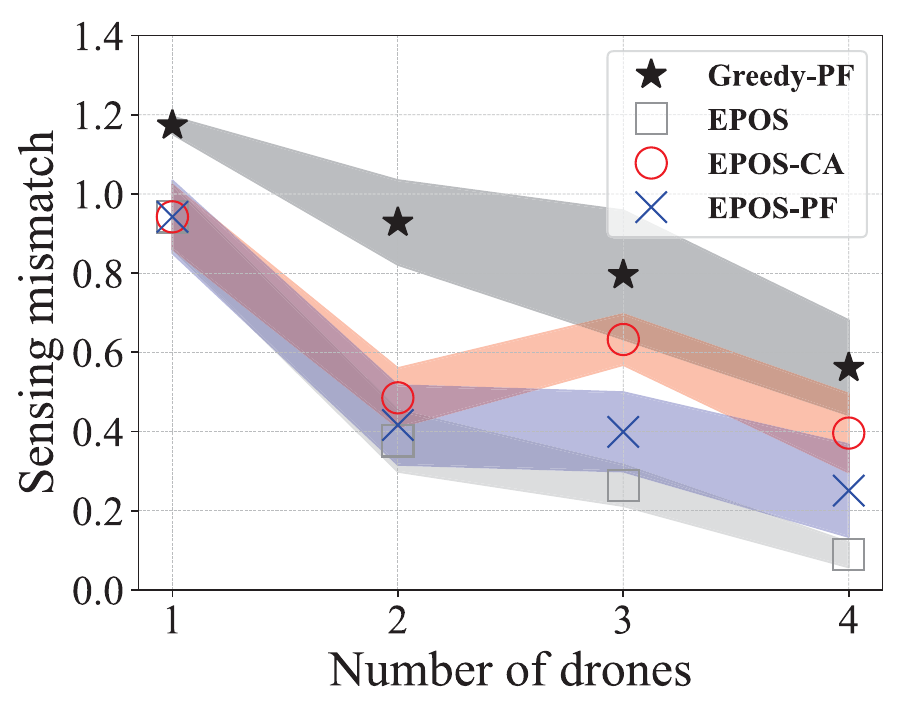}
		\label{fig:eval_mismatch}
	}
 	\caption{Energy, sensing and collision performance comparison of drones.\textbf{ \emph{EPOS-PF} significantly mitigates the risk of collisions and maintains low sensing mismatch, while consuming higher energy than other baseline methods with a high number of drones.} The points markers denote hardware results whereas the shadow represents the errors in software results.}
	\label{fig:testbed_collision_sensing}
\end{figure}

\subsection{Analysis of results}
To validate the applicability and realism of M-SET, we compare the energy consumption of drones estimated in software and the actual energy consumption in hardware. Fig.~\ref{fig:cf_voltage} illustrates the actual voltage of Crazyflies within the logging system of Crazyswarm where $4$ Crazyflies only hover for $250$ seconds. The voltage is recorded per second and is used to calculate the actual energy consumption based on battery capacity (250mAh) and expected flight time (7min). Fig.~\ref{fig:cf_energy} compares, for each drone, the actual energy consumption with model-based estimated the one calculated during the planning phase of M-SET, with only approximately $120.4$ joules error. This is because Crazyflies spend some additional flying time to calibrate between departure and landing. 

Different methods are compared with 4 number of drones in total energy consumption. We firstly run the software with $40$ different areas of the city centre~\cite{barmpounakis2020new}, and then choose one of the results (the average of all results) for the execution in hardware. Fig.~\ref{fig:eval_energy} shows that the disparity of energy consumption with and without collision avoidance rises as the number of drones increases, especially for artificial potential field. This is because each drone needs to detect and avoid several collisions within a small testbed, extending their traveling time.

Albeit a high energy consumption, \emph{EPOS-PF} significantly reduces the risk of collisions compared to other baseline methods, as shown in Fig.~\ref{fig:eval_risk}. It can detect various types of collisions that \emph{EPOS-CA} cannot (e.g., the case where three drones collide). The low risk of collisions in \emph{EPOS-PF} further proves the applicability of M-SET. Besides, since the highest number of collisions belongs to destination-occupied collisions, as shown in Fig.~\ref{fig:eval_collision_types}, drones with low priority in \emph{EPOS-CA} continuously wait and sense the same cell, resulting in over-sensing and under-sensing. This increases the sensing mismatch of \emph{EPOS-CA} compared to \emph{EPOS} as shown in Fig.~\ref{fig:eval_mismatch}. In contrast, \emph{EPOS-PF} dynamically repels and attracts drones to different areas, thereby mitigating over-sensing and under-sensing, and resulting in a sensing mismatch approximately $21.93\%$ lower than \emph{EPOS-CA}. \emph{Greedy-PF} chooses the navigation and sensing with only one cell to minimize the energy consumption, mitigating the risk of collision, but still confronts over-sensing and under-sensing.

In summary, several new insights on experimental results are listed as follows: (i) M-SET using \emph{EPOS-PF} improves sensing quality of traffic monitoring while avoiding collisions, making it energy-efficient with a limited number of drones. (ii) The accurate estimation of energy consumption and low risk of collisions validate M-SET as a proof-of-concept, proving its feasibility and safety in real-word applications. (iii) The expenditure of M-SET (each Crazyflie with necessary decks costs around $\$600$) is significantly lower than outdoor experimentation with larger drones with cameras (e.g., Phantom 4 Pro at around $\$1600$). (iv) M-SET eliminates concerns about licensing, atmospheric conditions, and privacy violations.

\section{Conclusion and Future work}
In conclusion, this paper introduces a novel testbed (M-SET) to integrate collision avoidance method to distributed sensing with drones exhibiting swarm intelligence. Designed for indoor lab environments, M-SET simplifies complex outdoor sensing scenarios  while enhancing the realism of experimentation. As a proof-of-concept, this paper demonstrates the applicability of a decentralized multi-agent collective learning algorithm~\cite{pournaras2018decentralized} and an artificial potential field algorithm~\cite{fedele2018obstacles} to coordinate drones navigation and sensing for traffic monitoring. The results highlight the potential of M-SET and provide valuable opportunities for the broader community to enhance drone control in low-cost, safe and efficient distributed sensing scenarios for Smart Cities. 

The testbed opens up several promising avenues for further improvements: (i) Using advanced hardware (e.g., ultrasonic sensor) for real-time collision avoidance. (ii) Incorporating different drones with different sensors and data collection capabilities. (iii) Implementing and integrating other swarm intelligence algorithms and reinforcement learning~\cite{hsu2020reinforcement} for enhanced online testbed operations.

\section*{Acknowledgment}
\footnotesize{
This research is supported by a UKRI Future Leaders Fellowship (MR-/W009560/1): \emph{Digitally Assisted Collective Governance of Smart City Commons–ARTIO}, and the European Union, under the Grant Agreement GA101081953 for the project H2OforAll—\emph{Innovative Integrated Tools and Technologies to Protect and Treat Drinking Water from Disinfection Byproducts (DBPs)}. Views and opinions expressed are, however, those of the author(s) only and do not necessarily reflect those of the European Union. Neither the European Union nor the granting authority can be held responsible for them. Funding for the work carried out by UK beneficiaries has been provided by UKRI under the UK government’s Horizon Europe funding guarantee [grant number 10043071]. Thanks to Emmanouil Barmpounakis and Nikolas Geroliminis for their support in using the pNEUMA dataset. 
}




\bibliographystyle{unsrt}
\bibliography{reference}

\end{document}